\documentclass{article}

\usepackage{arxiv}

\usepackage[utf8]{inputenc} 
\usepackage[T1]{fontenc}    
\usepackage{hyperref}       
\usepackage{url}            
\usepackage{booktabs}       
\usepackage{amsfonts}       
\usepackage{nicefrac}       
\usepackage{microtype}      
\usepackage{lipsum}		
\usepackage{graphicx}
\usepackage{doi}
\usepackage{comment}
\usepackage[font={small}]{caption}

\title{Using simulation to quantify the performance of automotive perception systems}

\author{{Zhenyi Liu}
        \thanks{Send correspondence to zhenyiliu@stanford.edu.} \\
	Stanford University\\
	Stanford, CA 94305, USA \\	
	\And
        Devesh Shah \\
	Ford Motor Company \\
        Dearborn, MI, 48124, USA\\
        \And
        Alireza Rahimpour  \\
	Ford Motor Company \\
        Palo Alto, CA 94304, USA\\
        \And
	Devesh Upadhyay \\
        Ford Motor Company \\
        Dearborn, MI, 48124, USA\\      
 	    \And
        Joyce Farrell \\
	Stanford University\\
	Stanford, CA 94305, USA \\
    \And
        Brian A Wandell \\
	Stanford University\\
	Stanford, CA 94305, USA \\
}

\renewcommand{\shorttitle}{\textit{arXiv} Template}

\hypersetup{
pdftitle={Designing scenes to quantify performance of automotive perception systems},
pdfsubject={q-bio.NC, q-bio.QM},
pdfauthor={Zhenyi Liu, Brian Wandell},
pdfkeywords={Keywords go here},
}

\begin{document} 

\pagestyle{empty} 
\setcounter{page}{301} 

\maketitle

\begin{abstract}
The design and evaluation of complex systems can benefit from a software simulation - sometimes called a digital twin. The simulation can be used to characterize system performance or to test its performance under conditions that are difficult to measure (e.g., nighttime for automotive perception systems). We describe the image system simulation software tools that we use to evaluate the performance of image systems for object (automobile) detection. We describe experiments with 13 different cameras with a variety of optics and pixel sizes. To measure the impact of camera spatial resolution, we designed a collection of driving scenes that had cars at many different distances. We quantified system performance by measuring average precision and we report a trend relating system resolution and object detection performance. We also quantified the large performance degradation under nighttime conditions, compared to daytime, for all cameras and a COCO pre-trained network.
\end{abstract}

\section{INTRODUCTION}
\label{sec:intro}  

 The field of imaging systems is expanding into new applications. Initially driven by rapid growth in the consumer photography market, the applications are now encompassing the automotive, robotics, and medical industries. The diverse range of applications and large markets will continue to drive innovation in the design and development of imaging systems.

An image system typically comprises optics, sensors, as well as task-specific image processing of the sensor data.  We are developing end-to-end simulation software to speed the design and evaluation of imaging systems. To carry out a simulation we must define the scene radiance and compute how the scene is transformed by the imaging system (optics, sensor) to create simulated camera images. These images are then processed for a specific goal, for example by submitting the images to neural networks trained to detect objects in the camera images or alternatively for processing by an image systems pipeline for display rendering and viewing.

This paper describes methods that use image systems simulation to evaluate  performance for object detection in an automotive application. We begin by defining a collection of scenes that are used to test performance. We then assess system performance in the object detection task, measuring average precision (similar to d-prime). In the examples presented here, each scene in the collection includes an object we aim to detect (a car) at one of a range of distances. The collections we create are used to assess system performance under different scene conditions (day or night), or different camera properties (optics and sensor resolution). Using standardized scene collections enables us to make meaningful comparisons between different imaging systems and different ambient conditions.

The approach we present expands upon conventional image system performance evaluation; in common practice, a test target is defined and performance is measured in the acquired image \cite{IEEE-p2020-pt}. It is expected that the system task - in this case object detection performance - can be predicted by the quality of the acquired image. End-to-end image systems simulations enables us to quantify the expected performance of the imaging system with respect to the object detection task.
\vspace{1mm}
\subsection{Related work}
Our work touches on three areas that are current topics in the imaging industry. 

First, image systems simulations (digital twins) are being actively pursued in academia and by commercial entities. For example, Nvidia~\cite{omniversevalidation-Kamel2021}, Tesla, and Waymo \cite{waymosimulator} all describe driving simulators that include sensors. The open-source CARLA simulator \cite{Dosovitskiy2017-if} is widely used to create  useful labeled synthetic data of complex driving scenes including various types of motion. Simulations of perception and decision systems are used in robotics for training \cite{Choi2021-yl}. The identified goals include network training but also using the simulations as an accelerated, safe, and fully controlled virtual testing and verification environment.  

The system we are developing is designed to simulate and evaluate the image system components, optics, sensors, and acquisition policies. This emphasis on quantitative analysis of these components is complementary to the emphasis in other simulators, which begin with 8-bit RGB images, often from unspecified devices, with little attention to the optics and sensor physics.

Second, there is an effort to quantify image quality of devices used in automotive applications. The IEEE P2020 committee  on automotive image quality \cite{IEEE-p2020-pt}is one example. There is a growing consensus, supported by the analysis in this paper and others at the EI 2023 conference \cite{MullerEI2023}, that the correlation between the standard measures of image quality (MTF\textsubscript{50}) and object detection performance is imperfect. 

Third, the difficulty in achieving high quality simulations and network training for nighttime driving images has been recognized in the literature \cite{Wu2019-night-rf,Sun2019-cy, Schutera2021-io}. This is an important area for simulation, because it is very difficult, perhaps impossible, to accurately label nighttime driving scenes. The big impact of the low illumination levels and high dynamic range found in nighttime driving make it imperative to design new approaches for image systems under night time conditions. This includes extending the simulations to accurately represent the nighttime scenes and adding improved simulations of camera flare, which is important but difficult to model component of the image system \cite{Dai2022-flare-ox,Wu2020-flare-bl}.
\section{Methods}

\subsection{Image systems simulation}

\begin{figure} [t]
    \centering
    \includegraphics[width=1\linewidth]{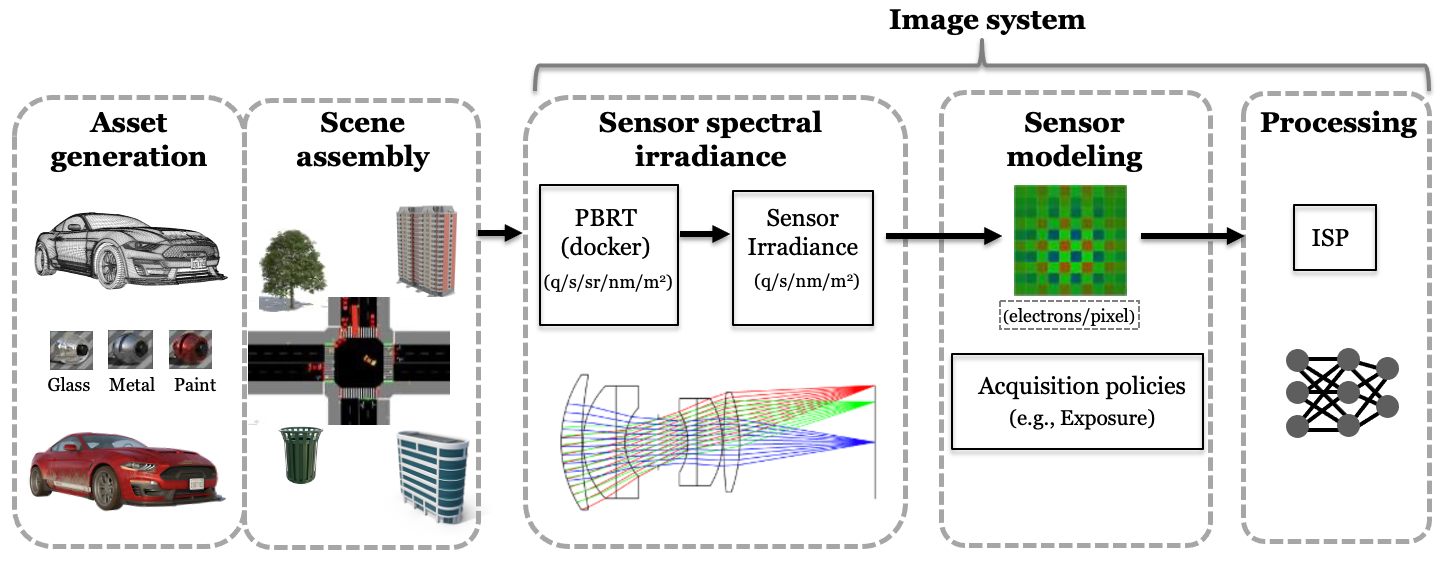}
    \caption{The image system simulation pipeline. We define scenes from assets, their spatial layout, and the lighting. Computer graphics renders these scenes through an optics model into the sensor spectral irradiance. Using a sensor model, the spectral irradiance is converted into image data that is input to the network for processing.  The optics, sensor, and processing (e.g., demosaicking, color balancing, neural networks) comprise the image system.  The parameters are chosen to match the simulated system.}
    \label{fig:simulationPipeline}
\end{figure} 

Figure \ref{fig:simulationPipeline} is an overview of the simulation pipeline we used for evaluating automobile detection. The pipeline begins by defining a scene (spectral radiance) and ends with the neural network estimate of the bounding box of a car. The scene is created by selecting assets (e.g., roads, cars, trees, pedestrians) from a database and assembling them into a spatial layout on the road. Each asset is represented by a mesh with material properties (e.g., metals, glass, cloth, leaves) are defined by rendering software parameters. Lighting is a critical component of the scene; the rendering software has multiple ways to define environmental lighting (sky maps) and area lights (headlamps, street lights). For this study, camera and object motion were not simulated.  

We then rendered the assembled scene through an optics model, calculating the expected spectral irradiance at the sensor surface. The rendering software is a modified version of the open-source Physically Based Ray Tracing (PBRT version 4) \cite{pharr-book}. The PBRT software models multi-element lenses, including spherical or bi-convex designs, including materials with an index of refraction that can vary with wavelength \cite{iset3dsoftware}.  The software has been containerized (Docker) to simplify sharing.

 Using the ISETCam software\cite{Farrell2003-rb,Farrell2008-sc,Farrell2012-ma,isetcamsoftware}, we convert the sensor spectral irradiance into electrons captured at each pixel, which are then converted into digital values. ISETCam simulates multiple sensor properties including photon noise, electrical noise, color filter arrays, and sensor geometry. We chose the parameters of the sensor to match the Sony IMX363, a widely used sensor. The output at this stage of the simulation is a synthetic camera image whose values closely resembles the values measured from a real camera, as validated in an empirical study\cite{Lyu2022-kg}. The digital values of the synthetic camera images also depend on camera modules that (a) determine the acquisition parameters, such as exposure duration and focus, and (b) the image processing techniques, such as demosaicking and color transformations. 

Finally, we submitted the digital images to an object detection network (YOLOv5) which was pre-trained to identify and locate the images from the Common Objects in Context (COCO) data set \cite{cocodataset}. We used the methods defined by COCO to calculate the average precision for detecting cars \cite{cocodataset}; briefly, the average precision is by comparing the estimated bounding boxes with the ground truth, using overlap (intersections of unions) for and range of thresholds 0.5:0.05:0.95.

\begin{figure}[b]
    \centering
    \includegraphics[width=1\linewidth]{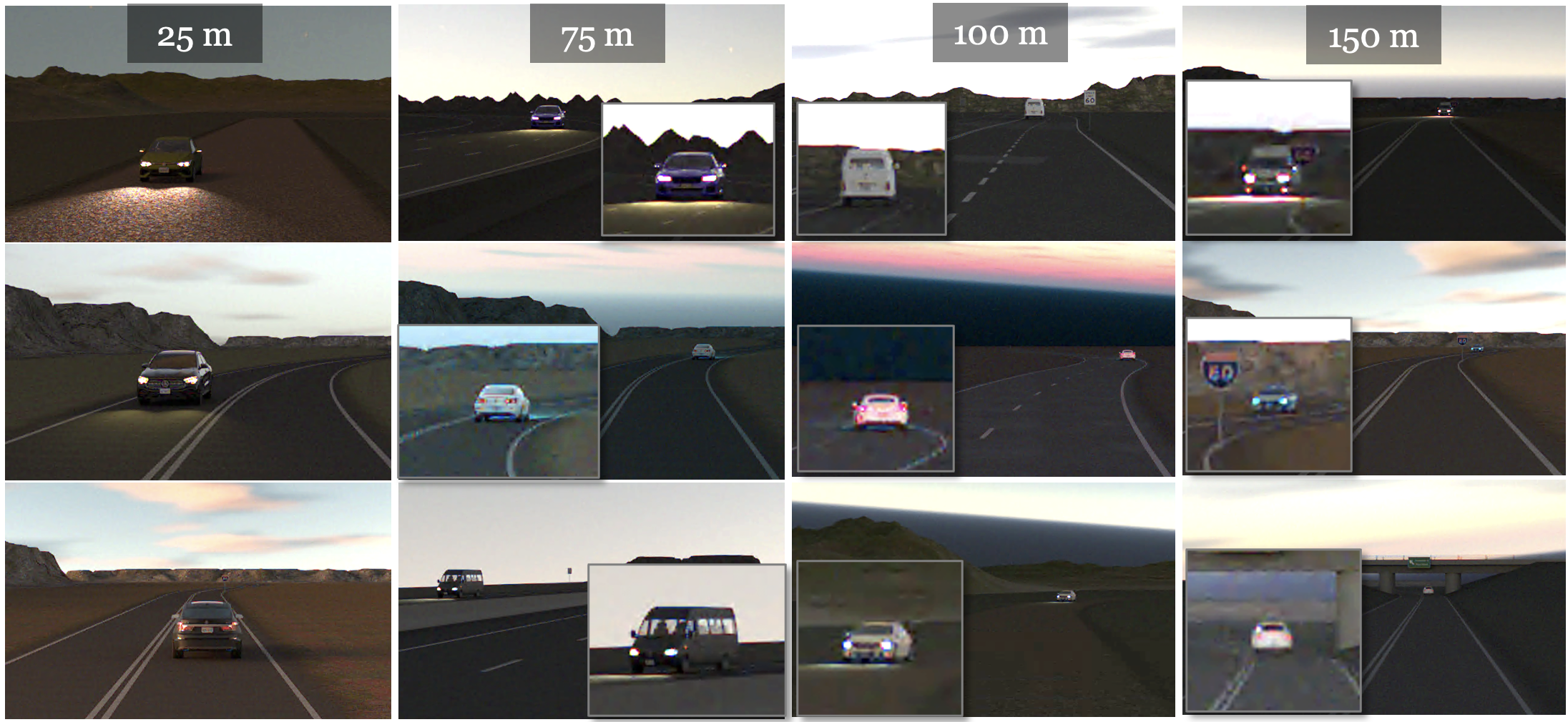}
    \caption{Examples from the scene collection for assessing spatial resolution. Each column shows three of the 50 scenes comprising a randomly selected car on a rural road.  In some of the images we use an inset to render the car. The scenes can be rendered under different illumination conditions, day to dusk are illustrated. We did not include streetlights in this collection.}
    \label{fig:metricScenes}
\end{figure}

\subsection{Metric scenes}
We design scene collections to probe specific aspects of the system performance. To measure the effective spatial resolution, we created a scene collection of single cars on a rural road.  These were created using a library of 3D assets that contains 50 high-resolution cars, 20 roads (400 m length, 200 m width), street lights, trees, and 30 high resolution HDR skymaps. To assemble a scene, we randomly select a car, a road and a skymap from the library. We created 300 scenes (spectral radiance data), each with a single car placed at one of six different distances from the simulated camera (25m, 50m, 75m, 100m, 150m, and 200m); there are 50 scenes for each distance. The car assets have headlights and brake lights, and the scenes have ambient illumination and streetlights. We can simulate the same scene geometry over a broad range of ambient illumination conditions, from daytime to nighttime.

We process these scenes through the image system simulation, calculating the bounding box when the car is detected. We quantify detection performance using the intersection of union (IoU) for the bounding box of the car and the estimate from YOLO. Then, we calculate the average precision (AP) of the image system at each distance using the AP calculation defined by COCO\cite{cocodataset}.

\subsection{Cameras}
We are particularly interested in the impact of the camera on system performance. For this study aimed at illustrating the method, we modeled cameras with diffraction-limited optics and Sony IMX363 sensors. The sensor spectral quantum efficiency, geometry, and electrical properties are described in a previous publication\cite{Lyu2022-kg}. In addition to the sensor, we must establish the acquisition policy, which describes how the camera module controls focus and exposure duration. We kept the focus at infinity, and adjusted a single exposure time in order to produce a sensor image that has peak voltage that is 90\% of the maximum (voltage swing) in the central region of the image. We limited the exposure duration to a maximum time of 16 ms (60 frames per second).

\begin{figure}[h]
    \begin{center}
    \includegraphics[width=0.8\linewidth]{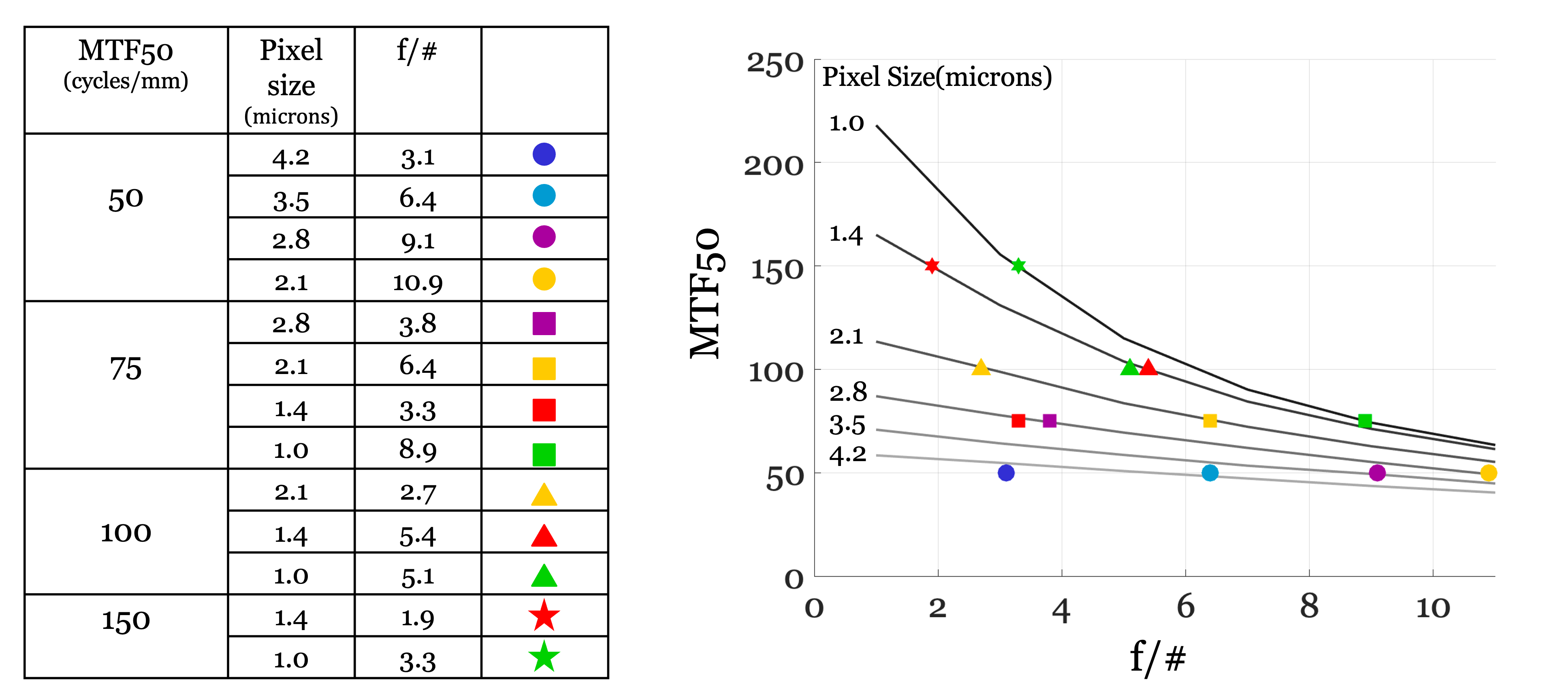}
    \end{center}
    \caption{Camera parameter for 13 different simulated imaging systems. The pixel size and f \# for each camera is listed in the table on the left. The MTF\textsubscript{50} value for each camera is also included in the table, as well as the symbols that are used to represent each camera. The graph in the right plots the MTF50 values as a function of pixel size and f/\#. }
    \label{fig:cameraParameters}
\end{figure}

In the experiments described below. We vary the f-number of the optics and the pixel size of the sensor (see Figure \ref{fig:cameraParameters}A), simulating 13 different imaging systems. These camera parameters combine to influence system resolution, and their combined effect is often characterized by the system modulation transfer function (MTF). This function describes how the system reduces image contrast over a range of spatial frequencies. The MTF\textsubscript{50} summarizes the whole MTF in a single number: the spatial frequency (cycles/mm) at which image contrast is reduced by half (50\%).

The MTF\textsubscript{50} is of particular interest because it is a widely used summary of camera resolution that is being considered as a standard by the IEEE P2020 Committee \cite{IEEE-p2020-pt}. We selected pixel size and optics parameters specifically to create cameras that (a) span a range of MTF\textsubscript{50} values, and (b) achieve the same MTF\textsubscript{50} with different combinations of f-number and pixel size (Figure \ref{fig:cameraParameters}B). The expectation is that cameras with equal MTF\textsubscript{50} will have equal system spatial resolution.

\section{RESULTS}

\subsection{Quantifying image system performance}
\begin{figure}[h]
    \centering
    \includegraphics[width=0.8\linewidth]{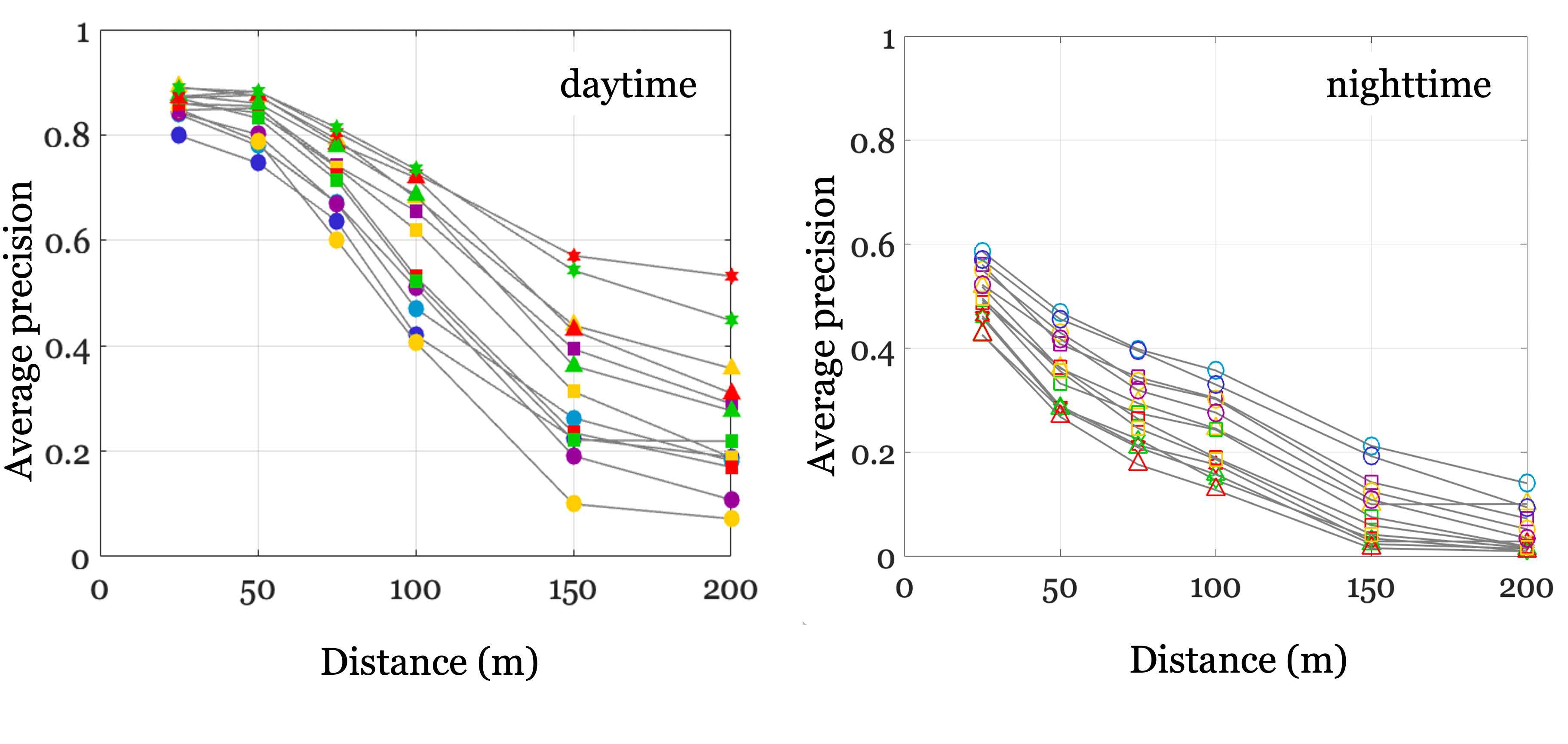}
    \caption{The average precision of each imaging system is plotted as a function of distance to the car for daylight (left, mean illuminance around 10-200 lux) and for nighttime (right, mean illuminance around 0.1-1 lux) illumination conditions.}
    \label{fig:avPrecisionVer2}
\end{figure}

For each of the 13 imaging systems, we calculated the average precision for detecting the cars as a function of distance to the car (Figure~\ref{fig:avPrecisionVer2}).  These are plotted separately for daylight (A) and for nighttime (B) illumination conditions. In all cases, system performance is relatively high for nearby cars and falls off with distance, as expected. The rate of performance degradation with increasing distance depends on the pixel size and f-number of the system, and the illumination condition (day vs. night) has a very large impact on performance. 

\subsection{The System Performance Map (SPM)}
\begin{figure}[h]
    \centering
    \includegraphics[width=0.85\linewidth]{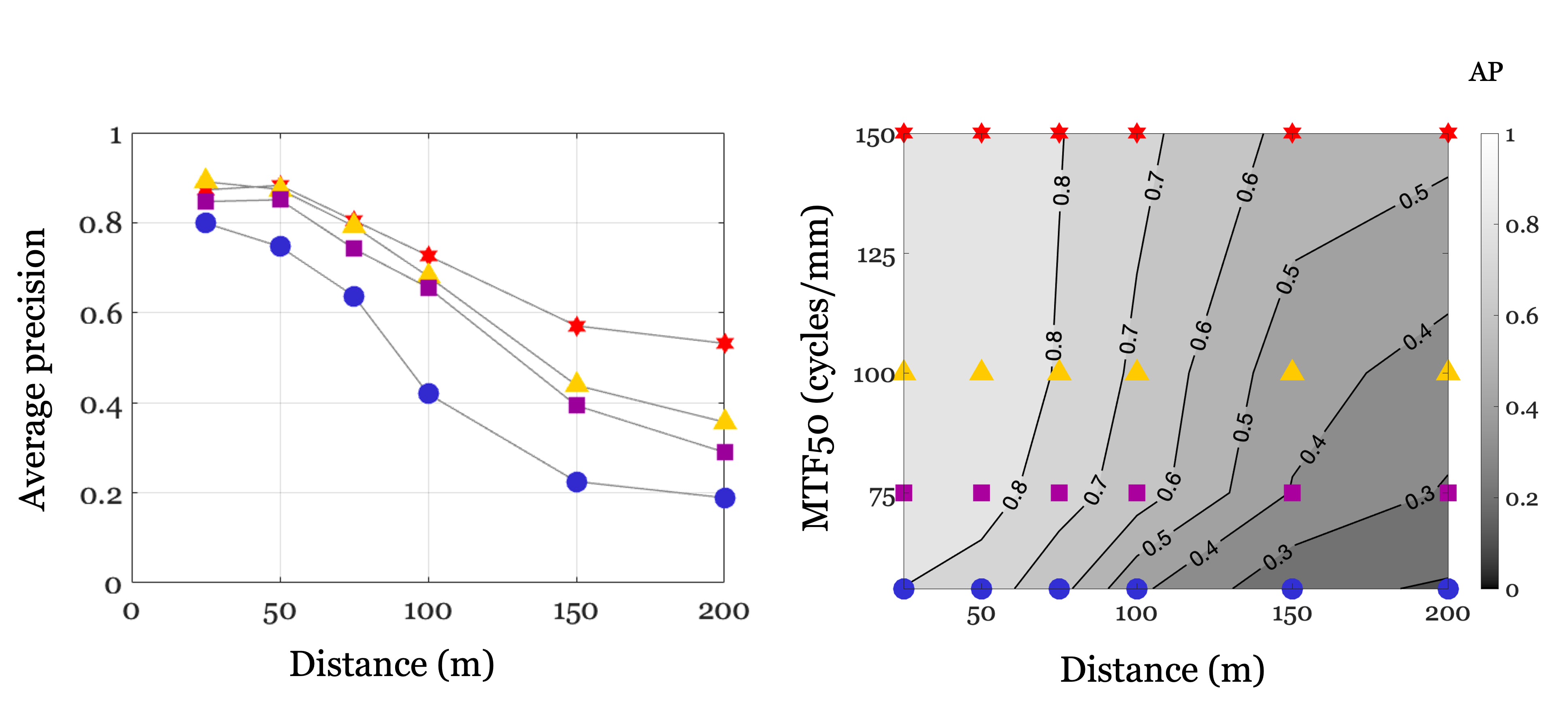}
    \caption{Calculating the System Performance Map (SPM). The figure on the left shows the average precision for four cameras with different pixel size and f/\#, resulting in four different MTF\textsubscript{50} values (refer to Figure \ref{fig:cameraParameters} for the values). The image on the right indicates how we create an image matrix with the average precision as a function of distance for these four cameras. The specific values are entered into the matrix, and we then use a contour plotting to interpolate the average precision levels (gray shading). The iso-performance curves are also shown.}
    \label{fig:SPMExplain}
\end{figure}

We use contour plots, referred to as System Performance Maps (SPMs), to visualize the system performance. Figure \ref{fig:SPMExplain} illustrates how to create an SPM. In this example, we selected four imaging systems with four different MTF\textsubscript{50} values.  The average precision curves for these four systems are on the left. We then create a matrix in which the average precision is entered as a function of MTF\textsubscript{50} (rows) and object distance (columns).  These are represented by the colored symbols in the image on the right.

\begin{figure}[t]
    \centering
    \includegraphics[width=0.84\linewidth] {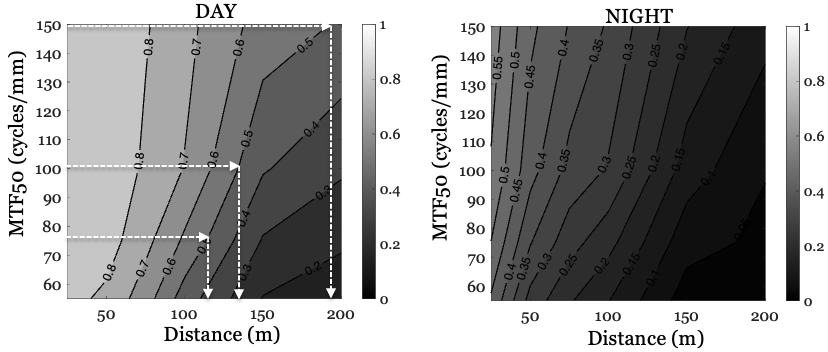}
    \caption{System Performance Tradeoff Maps (SPMs) calculated by combining all 13 cameras. Average precision is plotted as a function of MTF\textsubscript{50} and object distance for daylight (left) and nighttime (right) driving conditions. The dashed lines in the left figure points to the object distance where detection falls to an average precision of $0.50$ (referred to as the OD\textsubscript{50}) for a camera that has an MTF\textsubscript{50} of 100 cycles/mm.}
    \label{fig:SPM_MTF50}
\end{figure}

We calculated SPMs using the data from all 13 cameras.  Because the performance level is very different under high and low light conditions, we made two different SPMs (Figure \ref{fig:SPM_MTF50}).  In both cases, the average precision decreases with object distance, and it increases with MTF\textsubscript{50}. The size of these differences are small compared to the impact of light level.

The acquisition policy and additional network training are both likely to be important factors in improving the performance under night driving conditions. We are now using simulation to understand the impact of the acquisition policies and training in order to improve performance under night time conditions. 

\subsection{The MTF\textsubscript{50} and OD\textsubscript{50}}
\begin{figure}[th!]
\centering
    \includegraphics[width=0.5\linewidth]{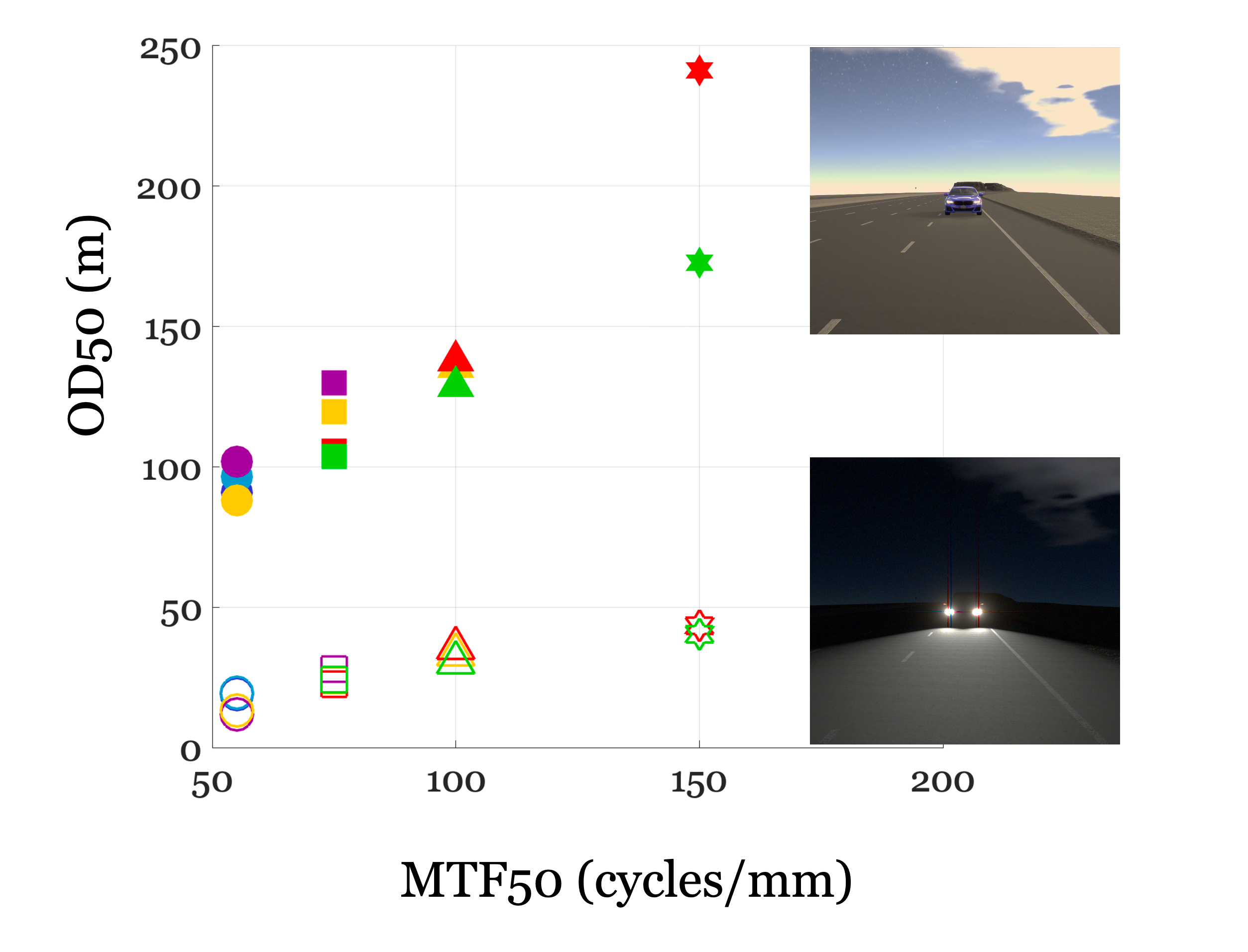}
    \caption{The dependence of object detection camera MTF\textsubscript{50}. We show the OD\textsubscript{50} for each of the thirteen camera models as a function of that camera's MTF\textsubscript{50} (see camera parameters in Figure \ref{fig:cameraParameters}). Solid and open symbols plot the OD\textsubscript{50} for daytime and nighttime conditions, respectively. The outlier (red filled star) arises because the average precision (Figure \ref{fig:avPrecisionVer2} for that camera never crosses the 0.50 level and the extrapolation is very poor. In a bootstrap analysis, we find that one standard deviation of the OD\textsubscript{50} estimate is on the order of 10 meters. Consequently the day and night estimates are very different, and there may be differences even among the daytime simulations with an equal MTF\textsubscript{50} (e.g., 75 cycles/mm).} 
    \label{fig:OD50vsMTF50}
\end{figure}

To further quantify the relationship between MTF\textsubscript{50} and spatial resolution, we measured the distance where the average precision for object detection is $0.50$ (OD\textsubscript{50}). Figure \ref{fig:OD50vsMTF50}A shows the relationship between each system's MTF\textsubscript{50} and its OD\textsubscript{50}.  The data are plotted separately for daytime and nighttime scenes.

The relationship is roughly linear, but the slope of OD\textsubscript{50} is higher under daytime compared to nighttime conditions. Thus, OD\textsubscript{50} increases more rapidly with MTF\textsubscript{50} when there is more ambient light; camera spatial resolution is not as significant a limiting factor under nighttime conditions.  Other factors, such as camera dynamic range or acquisition policies, are more important limitations. It might be better, for example, to increase well capacity and increase pixel size to achieve better performance under nighttime conditions.

\subsection{SPM and scene luminance}
To explore the relationship between performance and scene illumination level more fully, we simulated image system performance for a large range of scene illumination levels, a sensor pixel size of 1.4 micron, and an f/\# of 2.4.
  
\begin{figure}[h!]
    \centering
    \includegraphics[width=0.7\linewidth]{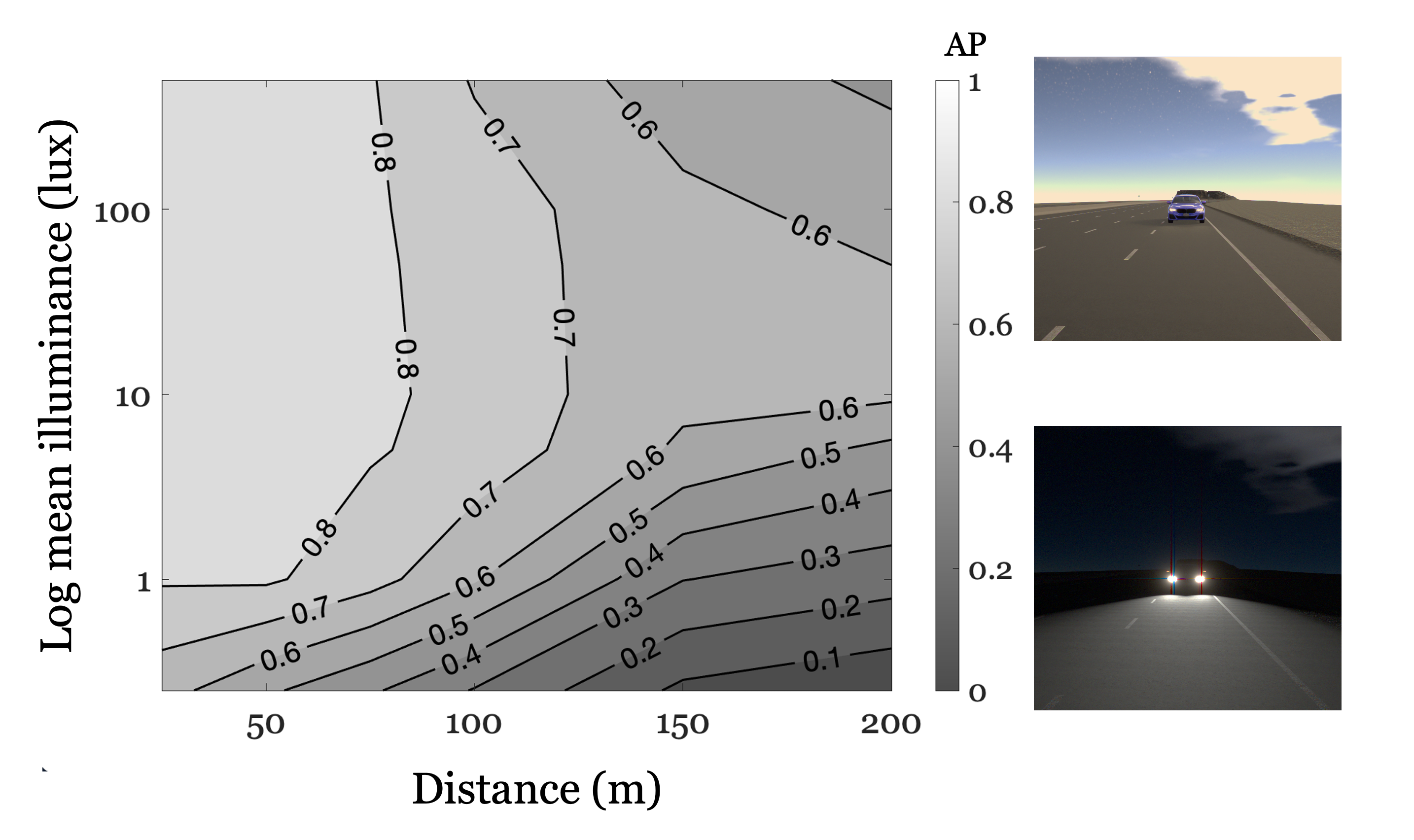}
    \caption{System performance map at different scene luminance levels. We simulated a camera with 140 MTF\textsubscript{50}.}
    \label{fig:SPMLuminance}
\end{figure}

Figure \ref{fig:SPMLuminance} shows that at low performance levels (average precision $<$ 0.6), object detection rises significantly with luminance level. But when scene luminance reaches 10 lux, object detection is not impacted by increasing illuminance. In exploring the non-monotonic performance a long distances (150m), we found that the sensor images were badly exposed.  Specifically, the acquisition policy created some images that were saturated and this limited performance for detecting these cars.

\section{DISCUSSION}
A complex array of factors determines the ability of an image system to detect objects. The scene, camera and neural network can all be performance-limiting factors. The ability to diagnose the limits on performance can be greatly improved by trustworthy image systems simulation. This paper describes and illustrates open-source tools we are developing to perform such simulations and analyze system performance for a driving application \cite{liu2019-softprototyping, Liu2020-generalization,Liu2021-depth-radiance}. We draw the reader's attention to three main ideas.

First, assessing system performance can benefit from carefully designed collections of test scenes. This paper measures the performance for a simple collection of driving scenes (a single car on a rural road) that is designed to quantify spatial resolution. We then modified these scenes, systematically varying the scene luminance and adding headlights to the cars, and compared performance under nighttime and daytime conditions. Defining a collection of scenes is analogous to defining a spatial resolution target, such as the slanted edge used to define a modulation transfer function in conventional image quality assessment.

Second, we illustrated how to compare the performance of a collection of cameras on the same scenes. Because the spatial resolution of the scenes is higher than the resolution of the cameras, we could use the same set of scenes as we assessed the cameras with a range of pixel sizes and f-numbers.  This exploration revealed a trend relating camera spatial resolution (MTF\textsubscript{50}) and object detection (Figure~\ref{fig:OD50vsMTF50}). We note, however, that even within this narrow range of cameras, the object detection scores differed by as much as three standard deviations when the MTF\textsubscript{50} was equal. A change in the MTF\textsubscript{50} (from 75 cyc/mm to 125 cy/mm) is very significant for consumer photography. The impact on automobile detection, however, is an increase of about 20 meters (Figure \ref{fig:SPM_MTF50}), and there are reliable differences in the object detection performance between systems with the same MTF\textsubscript{50} (Figure \ref{fig:OD50vsMTF50}).  Far larger effects are caused by variations in the illumination conditions.

Third, the simulations demonstrate the importance of environmental lighting conditions (day vs. night). The largest impact on object detection is the scene illuminance level and dynamic range; the mean illuminance of the nighttime driving scenes is very low; the car and street lights are extremely bright. Other environmental factors are very likely to emerge as large effects through further exploration (rain and fog; smoke; scene complexity; street signs; surface materials; plants). Maintaining safe performance across variations in these conditions may require adjusting acquisition policies (exposure control and perhaps focus; perhaps directional control of the camera's field of view). Image systems simulations can be helpful in analyzing and designing such system features, enabling the co-design of many system components.

\section*{ACKNOWLEDGMENTS}       
We thank David Cardinal and Doug Ward for their help with software and computer infrastructure used in this project.  We also thank Krithin Kripakaran, Dylan Li and Michael Xu for their help in preparing the automobile assets. We thank Jiayue Xie and Chuxi Yang from Tsinghua University for their help in preparing the road scenes with roadrunner and building the 3d assets library for night time.

\bibliographystyle{ieeetr}
\bibliography{references.bib} 

\end{document}